\documentclass[11pt,a4paper]{article}
\usepackage[hyperref]{acl2019}
\usepackage{times}
\usepackage{latexsym}
\usepackage{algorithm}
\usepackage{algpseudocode}
\usepackage{algorithmicx}
\usepackage{url}
\usepackage{booktabs}
\usepackage{amsmath}
\usepackage{amsthm}
\usepackage{amssymb}
\usepackage{bbm}
\usepackage{multirow}
\usepackage{graphicx}
\usepackage{caption}
\usepackage{todonotes} % insert [disable] to disable all notes

\aclfinalcopy % Uncomment this line for the final submission
\def\aclpaperid{88} %  Enter the acl Paper ID here

\DeclareMathOperator*{\argmax}{arg\,max}
\title{DoubleTransfer at MEDIQA 2019: \\Multi-Source Transfer Learning for Natural Language Understanding \\in the Medical Domain}
\author{Yichong Xu$^1$, Xiaodong Liu$^{2}$, Chunyuan Li$^{2}$, Hoifung Poon$^{2}$ and Jianfeng Gao$^{2}$ \\
	$^1$ Carnegie Mellon University\\
	$^2$ Microsoft Research\\
	\texttt{yichongx@cs.cmu.edu}\\
	\texttt{\{xiaodl, Chunyuan.Li, hoifung, jfgao\}@microsoft.com }
}
\begin{document}
		\maketitle
        \begin{abstract}
            This paper describes our competing system to enter the MEDIQA-2019 competition. We use a multi-source transfer learning approach to transfer the knowledge from MT-DNN \cite{liu2019multi} and SciBERT \cite{beltagy2019scibert} to natural language understanding tasks in the medical domain. For transfer learning fine-tuning, we use multi-task learning on NLI, RQE and QA tasks on general and medical domains to improve performance. The proposed methods are proved effective for natural language understanding in the medical domain, and we rank the first place on the QA task.
        \end{abstract}
        
        \section{Background}
        The MEDIQA 2019 shared tasks \cite{abacha2019multi} aim to improve the current state-of-the-art systems for textual inference, question entailment and question answering in the medical domain. This ACL-BioNLP 2019 shared task is motivated by a need to develop relevant methods, techniques and gold standards for inference and entailment in the medical domain and their application to improve domain-specific information retrieval and question answering systems. The shared task consists of three parts: i) natural language inference (NLI) on MedNLI, ii) Recognizing Question Entailment (RQE), and iii) Question Answering (QA).
        
        Recent advancement in NLP such as BERT \cite{devlin2018bert} has facilitated great improvements in many Natural Language Understanding (NLU) tasks~\cite{liu2019multi}. BERT first trains a language model on an unsupervised large-scale corpus, and then the pretrained model is fine-tuned to adapt to downstream NLU tasks. This fine-tuning process can be seen as a form of transfer learning, where BERT learns knowledge from the large-scale corpus and transfer it to downstream tasks. 
        
        We investigate NLU in the medical (scientific) domain. From BERT, we need to adapt to i) The change from general domain corpus to scientific language; ii) The change from low-level language model tasks to complex NLU tasks. Although there is limited training data in NLU in the medical domain, we fortunately have pre-trained models from two intermediate steps:
        \begin{itemize}
            \item General NLU embeddings: We use MT-DNN \cite{liu2019multi} trained on GLUE benchmark\cite{wang2019glue}. MT-DNN is trained on 10 tasks including NLI, question equivalence, and machine comprehension. These tasks correspond well to the target MEDIQA tasks but in different domains.
            
            \item Scientific embeddings: We use SciBERT \cite{beltagy2019scibert}, which is a BERT model, but trained on SemanticScholar scientific papers. Although SciBERT obtained state-of-the-art results on several single-sentence tasks, it lacks knowledge from other NLU tasks such as GLUE.
            %\item NLU in general domain: We use MT-DNN \cite{liu2019multi} trained on GLUE benchmark\cite{wang2019glue}. MT-DNN is trained on 10 tasks including NLI, question equivalence, and machine comprehension. These tasks correspond well to the target MEDIQA tasks but in different domains.
            %\item Language model in the scientific domain: We use SciBERT \cite{beltagy2019scibert}, a model with the same architecture as BERT, but trained on SemanticScholar scientific papers. Although SciBERT obtained state-of-the-art results on several single-sentence tasks, it lacks knowledge from other NLU tasks such as GLUE.
        \end{itemize}
        In this paper, we investigate different methods to combine and transfer the knowledge from the two different sources and illustrate our results on the MEDIQA shared task. We name our method as DoubleTransfer, since it transfers knowledge from two different sources. Our method is based on fine-tuning both MT-DNN and SciBERT using multi-task learning, which has demonstrated the efficiency of knowledge transformation \cite{caruana1997multitask,liu2015mtl, xu2018multi, liu2019multi}, and integrating models from both domains with ensembles.
        
        \begin{algorithm*}[htb!]
    \caption{Multi-task Fine-tuning with External Datasets}
    \begin{algorithmic}[1]        
        \Require{In-domain datasets $\mathcal{D}_1,...,\mathcal{D}_{K_1}$, External domain datasets $\mathcal{D}_{K_1+1},...,\mathcal{D}_{K_2}$,  max\_epoch, mixture ratio $\alpha$}
        \State Initialize the model $\mathcal{M}$ \label{step:init}
        \For{epoch$=1,2,...$, max\_epoch}
        \State Divide each dataset $\mathcal{D}_k$ into $N_k$ mini-batches $\mathcal{D}_k=\{b_1^k,...,b_{N_k}^k\}$, $1\leq k\leq K_2$ 
        \State $S\leftarrow \mathcal{D}_1\cup\mathcal{D}_2\cup\cdots\cup \mathcal{D}_{K_1}$
        \State $N\leftarrow N_1+N_2+\cdots+N_{K_1}$
        \State Randomly pick $\lfloor\alpha N \rfloor$ mini-batches from $\bigcup_{k=K_1}^{K_2} \mathcal{D}_k$ and add to $S$
        \State Assign mini-batches in $S$ in a random order to obtain a sequence $B=(b_1,...,b_L)$, where $L=N+\lfloor\alpha N\rfloor$
        \For{each mini-batch $b\in B$}
        \State Perform gradient update on $\mathcal{M}$ with loss $l(b)=\sum_{(s_1,s_2)\in b} l(s_1,s_2)$
        \EndFor
        \State Evaluate development set performance on $\mathcal{D}_1,...,\mathcal{D}_{K_1}$
        \EndFor
        \Ensure{Model with best evaluation performance % $\mathcal{M}^*$ 
        }
    \end{algorithmic}
    \label{algo:mtfinetune_ratio}
\end{algorithm*}

        \noindent\textbf{Related Works.} 
        % \section{Related Work}
        Transfer learning has been widely used in training models in the medical domain. For example, \citet{romanov2018lessons} leveraged the knowledge learned from SNLI to MedNLI; a transfer from general domain NLI to medical domain NLI. They also employed word embeddings trained on MIMIC-III medical notes, which can be seen as a language model in the scientific domain. SciBERT \cite{beltagy2019scibert} studies transferring knowledge from SciBERT pretrained model to single-sentence classification tasks. Our problem is unique because of the prohibitive cost to train BERT: Either BERT or SciBERT requires a very long time to train, so we only explore how to combine the existing embeddings from SciBERT or MT-DNN.
        Transfer learning is also widely used in other tasks of NLP, such as machine translation \cite{bahdanau2014neural} and machine reading comprehension \cite{xu2018multi}.

\begin{figure*}
    \centering
    \vspace{-1mm}
    {
     \includegraphics[scale=0.5]{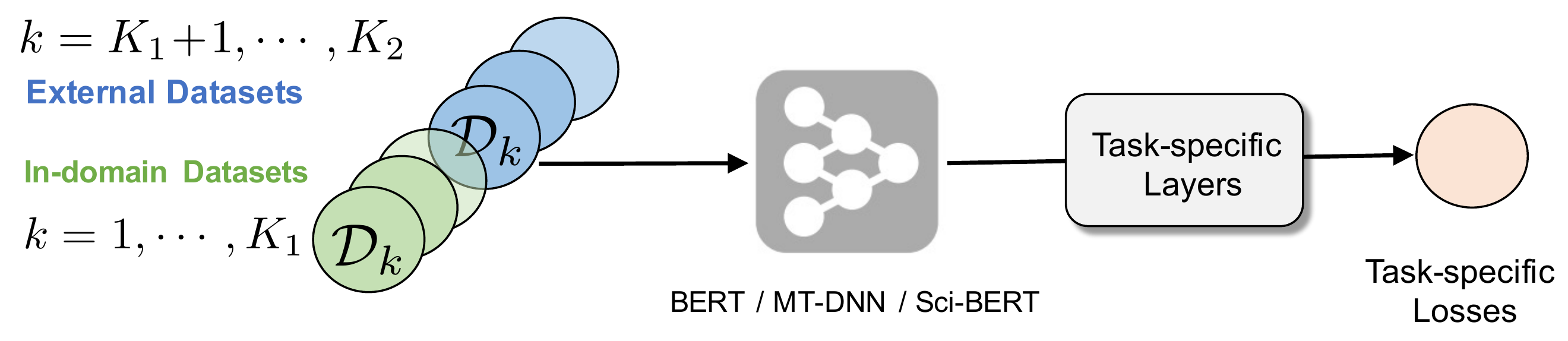}
     }
    \caption{Illustration of the proposed multi-source multi-task learning method.}
    \label{fig:msmt}
\end{figure*}

        \section{Methods}
        We propose a multi-task learning method for the medical domain data. It employs datasets/tasks from both medical domain and external domains, and leverage the pre-trained model such as MT-DNN and SciBERT for fine-tuning. An overview of the proposed method is illustrated in Figure~\ref{fig:msmt}. To further improve the performance, we propose to ensemble models trained from different initialization in the evaluation stage. Below we detail our methods for fine-tuning and ensembles.

        \subsection{Fine-tuning details}
        \textbf{Algorithm.} We fine-tune the two types of pre-trained models on all the three tasks using multi-task learning. As suggested by MEDIQA paper, we also fine-tune our model on MedQuAD \cite{abacha2019question}, a medical QA dataset. We will provide details for fine-tuning on these datasets in Section \ref{sec:detail}. We additionally regularize the model by also training on MNLI \cite{N18-1101}. To prevent the negative transfer from MNLI, we put a larger weight on MEDIQA data by sampling MNLI data with less probability. Our algorithm is presented in Algorithm \ref{algo:mtfinetune_ratio} and illustrated as Figure~\ref{fig:msmt}, which is a mixture ratio method for multi-task learning inspired by \citet{xu2018multi}.  We start with in-domain datasets $\mathcal{D}_1,...\mathcal{D}_{K_1}$ (i.e., the MEDIQA tasks, $K_1=3$) and external datasets $\mathcal{D}_{K_1+1},...,\mathcal{D}_{K_2}$ (in this case MNLI). We cast all the training samples as sentence pairs $(s_1,s_2)\in \mathcal{D}_k, k=1,2,...,K_2$. In each epoch of training, we use all mini-batches from in-domain data, while only a small proportion (controlled by $\alpha$ ) of mini-batches from external datasets are used to train the model. In our experiments, the mixture ratio $\alpha$ is set to 0.5. We use MedNLI, RQE, QA, and MedQuAD in medical domain as in-domain data and MNLI as external data. For MedNLI, we additionally find that using MedNLI as in-domain data and RQE, QA, MedQuAD as external data can also help boost performance. We use models trained using both setups of external data for ensembling.

        \noindent\textbf{Pre-trained Models.} We use three different types of initialization as the starting point for fine-tuning: i) the uncased MT-DNN large model from \citet{liu2019multi}, ii) the cased knowledge-distilled MT-DNN model from \citet{liu2019improving}, and iii) the uncased SciBERT model \cite{beltagy2019scibert}.
        % For model architecture, we use both cased and uncased model for MT-DNN from BERT-Large models and use the uncased model for SciBERT for initialization. 
        We add a simple softmax layer (or linear layer for QA and MedQuAD tasks) atop BERT as the answer module for fine-tuning. For initialization in step \ref{step:init} in Algorithm \ref{algo:mtfinetune_ratio}, we initialize all BERT weights with the pretrained weights, and randomly initialize the answer layers.    
        After multi-task fine-tuning, the joint model is further fine-tuned on each specific task to get better performance. We detail the training loss and fine-tuning process for each task in Section \ref{sec:detail}.
        
        \noindent\textbf{Objectives.} MedNLI and RQE are binary classification tasks, and we use a cross-entropy loss. Specifically, for a sentence pair $X$ we compute the loss
        \begin{equation*}
        \mathcal{L}(X)=-\sum_c \mathbbm{1}(X,c) \log(P_r(c|X)),
        \end{equation*}
        where $c$ iterates over all possible classes, $\mathbbm{1}(X,c)$ is the binary indicator (0 or 1) if class label $c$ is the correct classification for $X$, and $P_r(c|X)$ is the model prediction for probability of class $c$ for sample $X$. 
        
        %For QA and MedQuAD, 
        We formulate QA and MedQuAD as regression tasks, and thus a MSE loss is used. Specifically, for a question-answer pair $(Q,A)$ we compute the MSE loss as
        \[\mathcal{L}(Q,A)=( y - \mathtt{score}(Q,A))^2, \]
        where $y$ is the target relevance score for pair $(Q,A)$, and $\mathtt{score}(Q,A)$ is the model prediction for the same pair.

        \subsection{Model Ensembles}
        After fine-tuning, we ensemble models trained from MT-DNN and SciBERT, and using different setups of in-domain and external datasets. The traditional methods typically fuse models by averaging the prediction probability of different models. For our setting, the in-domain data is very limited and it tends to overfit; this means the predictions can be arbitrarily close to 1, favoring to more over-fitting models. To prevent over-fitting, we ensemble the models by using a majority vote on their predictions, and resolving ties using sum of prediction probabilities. Suppose we have $M$ models, and the $m$-th model predicts the answer $\hat{p}_m$ for a specific question. For the classification task (MedNLI and RQE), we have $\hat{p}_m\in \mathbb{R}^C$, where $C$ is the number of categories. Let $\hat{y}_m=\argmax_i \hat{p}_m^{(i)}$ be the prediction of model $m$, where $\hat{p}_m^{(i)}$ is the $i$-th dimension of $\hat{p}_m$. The final prediction is chosen as
        \[\hat{y}_{\text{ensemble}}=\argmax_{y\in \text{maj}(\{\hat{y}_m\}_{m=1}^M)}\sum_{m=1}^M \hat{p}_m^{(y)}. \]
        In other words, we first obtain the majority of predictions by computing the majority $\text{maj}(\{\hat{y}_m\}_{m=1}^M)$, and resolve the ties by computing the sum of prediction probabilities $\sum_{m=1}^M \hat{p}_m^{(y)}$. For QA tasks (QA and MedQuAD), the task is cast as a regression problem, where a positive number means correct answer, and negative otherwise. We have $\hat{p}_m\in \mathbb{R}$. We first compute the average score $\hat{p}_{\text{ensem}}=\frac{1}{M}\sum_{m=1}^M \hat{p}_m$. We also compute the prediction as $\hat{y}_m=I(\hat{p}_m\geq 0)$, where $I$ is the indicator function. We compute the ensemble prediction through a similar majority vote as the classification case:
        \[\hat{y}_{\text{ensem}}=\begin{cases}
        1, & \text{if}~ \sum_{m=1}^M \hat{y}_m>M/2\\
        0, & \text{if}~ \sum_{m=1}^M \hat{y}_m<M/2\\
        I(\hat{p}_{\text{ensem}}>0), & \text{otherwise.}
        \end{cases}\]
        To be precise, we predict the majority if a tie does not exist, or the sign of $\hat{p}_{\text{ensem}}$ otherwise. The final ranking of answers is carried out by first rank the (predicted) positive answers, and then the (predicted) negative answers.
    
        \subsection{Dataset-Specific Details \label{sec:detail}}
        \textbf{MedNLI:} Since the MEDIQA shared task uses a different test set than the original MedNLI dataset, we merge the original MedNLI development set into the training set and use evaluation performance on the original MedNLI test set. Furthermore, MedNLI and MNLI are the same NLI tasks, thus, we shared final-layer classifiers for these two tasks. For MedNLI, we find that each consecutive 3 samples in all the training set contain the same premise with different hypothesizes, and contains exactly 1 entail, 1 neutral and 1 contradiction. To the end, in our prediction, we constrain the three predictions to be one of each kind, and use the most likely prediction from the model prediction probabilities.\\
        \textbf{RQE:} We use the clinical question as the premise and question from FAQ as the hypothesis. We find that the test data distribution is quite different from the train data distribution. To mitigate this effect, we randomly shuffle half of the evaluation data into the training set and evaluate on the remaining half. \\
        \textbf{QA:} We use the answer as the premise and the question as the hypothesis. The QA task is cast as both a ranking task and a classification task. Each question is associated with a relevance score in $\{1,2,3,4\}$, and an additional rank over all the answers for a specific question is given. We use a modified score to incorporate both information: suppose there are $m$ questions with relevance score $s\in \{1,2,3,4\}$. Then the $i$-th most relevant answer in these $m$ questions get modified score $s-\frac{i-1}{m}$. In this way the scores are uniformly distributed in $(s-1,s]$. We shift all scores by $-2$ so that a positive score leads to a correct answer and vice versa. We also tried pairwise losses to incorporate the ranking but did not find it to boost the performance very much. 
        
        We find that the development set distribution is inconsistent with test data - the training and test set consist of both LiveQAMed and Alexa questions, whereas the development set seems to only contain LiveQAMed questions. We shuffle the training and development set to make them similar: We use the last 25 questions in original development set (LiveQAMed questions) and the last 25 Alexa questions (from the original training set) as our development set, and use the remaining questions as our training set.
        % So we shuffle a part of development data into the training set, and some Alexa questions into the development set\xiaodl{I didn't get it. Do you merge dev to training?}. 
        This results in 1,504 training pairs and 431 validation pairs. Due to the limited size of the QA dataset, we use cross-validation that divides all pairs into 5 slices and train 5 models by using each slice as a validation set. We train MT-DNN and SciBERT on both these 5 setups and obtain 10 models, and ensemble all the 10 models obtained. \\
        \textbf{MedQuAD:} We use 10,109 questions from MedQuAD because the remaining questions are not available due to copyright issues. The original MedQuAD dataset only contains positive question pairs. We add negative samples to the dataset by randomly sampling an answer from the same web page. For each positive QA pair, we add two negative samples. The resulting 30,327 pairs are randomly divided into 27,391 training pairs and 2,936 evaluation pairs. Then we use the same method as QA to train MedQuAD; we also share the same answer module between QA and MedQuAD.
       
\subsection{Implementation and Hyperparameters}
        We implement our method using PyTorch\footnote{\url{https://pytorch.org/}} and Pytorch-pretrained-BERT\footnote{\url{https://github.com/huggingface/pytorch-pretrained-BERT}}, as an extension to MT-DNN\footnote{\url{https://github.com/namisan/mt-dnn}}. We also use the pytorch-compatible SciBERT pretrained model provided by AllenNLP\footnote{\url{https://github.com/allenai/scibert}}. Each training example is pruned to at most 384 tokens for MT-DNN models and 512 tokens for SciBERT models. We use a batch size of 16 for MT-DNN, and 40 for SciBERT. For fine-tuning, we train the models for 20 epochs using a learning rate of $5\times10^{-5}$. After that, we further fine-tune the model from the best multi-task model for 6 epochs for each dataset, using a learning rate of $5\times 10^{-6}$. We ensemble all models with an accuracy larger than 87.7 for MedNLI, 83.5 for shuffled RQE, and 83.0 for QA. We ensemble 4 models for MedNLI, 14 models for RQE. For QA, we ensemble 10 models from cross-validation and 7 models using the normal training-validation approach.

        \section{Results}
        In this section, we provide the leaderboard performance and conduct an analysis of the effect of ensemble models from different sources.
        
        \subsection{Test Set Performance and LeaderBoards}
        % We summarize the leaderboard performance for each of the three tasks in this section. 
        The results for MedNLI dataset is summarized in Table \ref{tab:mednli-leaderboard}. Our method ends up the 3rd place on the leaderboard and substantially improving upon previous state-of-the-art (SOTA) methods.
        \begin{table}[tb!]
    \begin{center}
        \begin{tabular}{l | c c}
            \hline  Model & Dev Set & Test Set   \\ \hline
            %                    Our Models & Acc & Acc\\   \midrule
            %                    MT-DNN (Single)  & 88.61 & 89.90\\
            %                    SciBERT (Single) & 87.70 & 87.43\\
            %                    Ensemble & 91.71 & 93.84\\
            %                    \hline
            %                     & Acc & Acc\\\hline
            WTMed & - & \bf 98.0  \\
            PANLP & -  & 96.6  \\        
            \textbf{Ours} & \bf 91.7  & \bf93.8  \\
            Sieg & -  & 91.1  \\\hline    
            SOTA & 76.6 & - \\
            \hline
        \end{tabular}
    \end{center}
    \caption{\label{tab:mednli-leaderboard} 
        %                Our model performance and  We only submitted ensembles to the leaderboard, and the test performance was obtained after the competition. 
        The leaderboard for MedNLI task (\href{https://www.aicrowd.com/challenges/mediqa-2019-natural-language-inference-nli/leaderboards}{link}). Scores are accuracy(\%). Our method ranked the 3rd on the leaderboard. Previous SOTA method was from \cite{romanov2018lessons}, on the original MedNLI test set (used as dev set here).
    }
\end{table}

The results for RQE dataset is summarized in Table \ref{tab:rqe-leaderboard}. Our method ends up the 7th place on the leaderboard. Our method has a very large discrepancy between the dev set performance and test set performance. We think this is because the test set is quite different from dev set, and that the dev set is very small and easy to overfit to.

\begin{table}[tb!]
    \begin{center}
        \begin{tabular}{l | c c}
            \hline  Model & Dev Set & Test Set   \\ \hline
            PANLP & - & \bf 74.9  \\
            Sieg & -  & 70.6  \\    
            IIT-KGP & -  & 68.4  \\        
            \textbf{Ours} & \bf 91.7  & \bf66.2  \\
            \hline
        \end{tabular}
    \end{center}
    \caption{\label{tab:rqe-leaderboard} 
        The leaderboard for RQE task (\href{https://www.aicrowd.com/challenges/mediqa-2019-recognizing-question-entailment-rqe/leaderboards}{link}). Scores are accuracy(\%). Our method ranked the 7th on the leaderboard.
    }
\end{table}  

    The results for QA dataset is summarized in Table \ref{tab:qa-leaderboard}. Our method reaches the first place on the leaderboard based on accuracy and precision score and 3rd-highest MRR. We note that the Spearman score is not consistent with other scores in the leaderboard; actually, the Spearman score is computed just based on the predicted positive answers, and a method can get very high Spearman score by never predict positive labels.
    
\begin{table}[tb!]
    \begin{center}
        \begin{tabular}{@{\hskip1pt}l@{\hskip1pt}|@{\hskip1pt} c  c c c}
            \hline  Model & Acc & Spearman & Precision & MRR \\ \hline
            \textbf{Ours} & \bf 78.0  & 0.238 & \bf 81.91 & 0.937 \\
            PANLP & 77.7  & 0.180 & 78.1 & 0.938 \\
            Pentagon & 76.5  & 0.338 & 77.7 & \bf 0.962 \\
            DUT-BIM &74.5 & 0.106 & 74.7 & 0.906 \\
            \hline
        \end{tabular}
    \end{center}
    \caption{\label{tab:qa-leaderboard} 
        The leaderboard for QA task (\href{https://www.aicrowd.com/challenges/mediqa-2019-question-answering-qa/leaderboards}{link}). Our method ranked \#1 on the leaderboard in terms of Acc (accuracy). The Spearman score is not consistent with other scores in the leaderboard.
    }
\end{table}

    \subsection{Ensembles from Different Sources}
        We compare the effect of ensembling from different sources in Table \ref{tab:compare-ensemble}. We train 6 different models with different randomizations, with initializations from MT-DNN (\#1,\#2,\#3) and SciBERT (\#4, \#5,\#6) respectively. If we ensemble models with the same MT-DNN architecture, the resulting model only has around 1.5\% improvement in accuracy, compared to the numerical average of the ensemble model accuracies (\#1+\#2+\#3 and \#4+\#5+\#6 in Table \ref{tab:compare-ensemble}). On the other hand, if we ensemble three models from different sources (\#1+\#2+\#5 and \#1+\#5+\#6 in Table \ref{tab:compare-ensemble}), the resulting model gains more than 3\% in accuracy compared to the numerical average. This shows that ensembling from different sources has a great advantage than ensembling from single-source models.
    
    \begin{table}[h!]
    \begin{center}
    \begin{tabular}{@{\hskip1pt}l | c c}
        \hline  Model & Avg. Acc & Esm. Acc  \\ \hline
        \multicolumn{3}{c}{Single Model }  \\ \hline %\midrule
        \#1, MT-DNN & - & 88.61 \\
        \#2, MT-DNN & - & 88.33 \\
        \#3, MT-DNN & - & 87.84 \\
        \#4, SciBERT & - & 88.19 \\
        \#5, SciBERT & - & 87.70 \\
        \#6, SciBERT & - & 87.21 \\\hline \hline
        \multicolumn{3}{c}{Ensemble Model}  \\ \hline %\midrule
        \#1+\#2+\#3, MT-DNN & 88.26 & 89.7\\
        \#4+\#5+\#6, SciBERT & 87.70 & 89.2\\
        \#1+\#2+\#5, MultiSource & 88.21 & 91.6\\
        \#1+\#5+\#6, MultiSource & 87.84 & 90.4\\
        \#1-6, MultiSource & 87.98 & 91.3    \\    
        \hline
    \end{tabular}
\end{center}
\caption{\label{tab:compare-ensemble} 
    Comparison of ensembles from different sources. Avg.Acc stands for average accuracy, the numerical average of each individual model's accuracy. Esm.Acc stands for ensemble accuracy, the accuracy of the resulting ensemble model. For ensembles, MT-DNN means all the three models are from MT-DNN, and similarly for SciBERT; MultiSource denotes the ensemble models come from two different sources.
}
\end{table}    

\subsection{Single-Model Performance}
For completeness, we report the single-model performance on the MedNLI development set under various multi-task learning setups and initializations in Table \ref{tab:single-model}. 
(1) The {\em Na\"{i}ve} approach denotes only MedNLI, RQE, QA, MedQuAD is considered as in-domain data in Algorithm \ref{algo:mtfinetune_ratio} without any external data; 
(2)  The {\em Ratio} approach denotes that we consider MedNLI as in-domain data, and RQE, QA, MedQuAD as external data in Algorithm \ref{algo:mtfinetune_ratio}; 
(3)  The {\em Ratio+MNLI} approach denotes that we consider MedNLI, RQE, QA, MedQuAD as in-domain data and MNLI as external data in Algorithm \ref{algo:mtfinetune_ratio}. Note that MNLI is much larger than the medical datasets, so if we use RQE, QA, MedQuAD, MNLI as external data, the performance is very similar to the third setting. We did not conduct experiments on single-dataset settings, as previous works have suggested that multi-task learning can obtain much better results than single-task models \cite{liu2019multi,xu2018multi}.

Overall, the best results are achieved via using SciBERT as the pre-trained model, and multi-task learning with MNLI.  The models trained by mixing in-domain data (the second setup) is also competitive. We therefore use models from both setups for ensemble. 
\begin{table}[h!]
    \begin{center}
    \begin{tabular}{@{\hskip1pt}l | c c c}
        \hline  Init Model & Na\"{i}ve & Ratio & Ratio+MNLI\\ \hline
        MT-DNN & 86.9 & 86.2 & 87.8  \\
        MT-DNN-KD & 87.5 &88.2 & \bf 88.8\\
        SciBERT & 87.1 & 87.0 & \bf 89.4 \\
        \hline
    \end{tabular}
\end{center}
\caption{\label{tab:single-model} 
Single model performance on MedNLI developlment data. {\em Nai\"{i}ve} means simply integrating all medical-domain data; {\em Ratio} means using MedNLI as in-domain data and other medical domain data as external data; {\em Ratio+MNLI} means using medical domain data as in-domain and MNLI as external.
    % Comparison of ensembles from different sources. Avg.Acc stands for average accuracy, the numerical average of each individual model's accuracy. Esm.Acc stands for ensemble accuracy, the accuracy of the resulting ensemble model. For ensembles, MT-DNN means all the three models are from MT-DNN, and similarly for SciBERT; MultiSource denotes the ensemble models come from two different sources.
}
\end{table}
\section{Conclusion}
We present new methods for multi-source transfer learning for the medical domain. Our results show that ensembles from different sources can improve model performance much more greatly than ensembles from a single source. Our methods are proved effective in the MEDIQA2019 shared task.

		\bibliography{yichongref}
\bibliographystyle{acl_natbib}
	\end{document}